\documentclass[
]{ceurart}

\usepackage{listings}
\begin{document}

\copyrightyear{2025}
\copyrightclause{Copyright for this paper by its authors.
  Use permitted under Creative Commons License Attribution 4.0
  International (CC BY 4.0).}

\conference{CLEF 2025 Working Notes, 9 -- 12 September 2025, Madrid, Spain}

\title{mdok of KInIT: Robustly Fine-tuned LLM for Binary and Multiclass AI-Generated Text Detection}
\title[mode=sub]{Notebook for the PAN Lab at CLEF 2025}

\author[]{Dominik Macko}[%
email=dominik.macko@kinit.sk,
]
\address[]{Kempelen Institute of Intelligent Technologies, Bratislava, Slovakia}


\begin{abstract}
 The large language models (LLMs) are able to generate high-quality texts in multiple languages. Such texts are often not recognizable by humans as generated, and therefore present a potential of LLMs for misuse (e.g., plagiarism, spams, disinformation spreading). An automated detection is able to assist humans to indicate the machine-generated texts; however, its robustness to out-of-distribution data is still challenging. This notebook describes our \textbf{mdok} approach in robust detection, based on fine-tuning smaller LLMs for text classification. It is applied to both subtasks of Voight-Kampff Generative AI Detection 2025, providing remarkable performance (\textbf{1st rank}) in both, the binary detection as well as the multiclass classification of various cases of human-AI collaboration.
\end{abstract}


\maketitle

\section{Introduction}

Continuously increasing quality of texts generated by artificial intelligence (AI) technology, such as large language models (LLMs), causes that humans are no longer able to differentiate between human-written and high-quality machine-generated texts. Naturally, this arises concerns about potential LLM misuse, e.g., for accelerated generation of disinformation~\cite{vykopal-etal-2024-disinformation,zugecova2024evaluationllmvulnerabilitiesmisused}, plagiarism~\cite{wahle-etal-2022-large}, or frauds for academic exams~\cite{openai2023gpt4}. The automated means, also utilizing AI technology, are able to help humans to differentiate such texts; however, the automated detection performance is also not perfect (errors occur, such as false positives or false negatives). Further challenges are in application of the trained detectors to out-of-distribution data, i.e. data that are significantly different to data used for training (surprise data).

The \textit{Voight-Kampff Generative AI Detection 2025} shared task~\cite{bevendorff:2025b}, as a part of \textit{PAN} lab~\cite{bevendorff:2025a} at the \textit{CLEF 2025} conference, addresses two challenges in the machine-generated text detection problem area in two subtasks. Subtask~1 is focused on classical binary detection task distinguishing between machine-authored (class~1) and human-authored (class~0) texts, which addresses also the robustness of detectors to obfuscation and other surprise data. Subtask~2 is focused on human-AI collaborative text classification, covering six classes of collaboration, namely class~0 representing fully human-written texts; class~1 representing human-written, then machine-polished texts, class~2 representing machine-written, then machine-humanized texts, class~3 representing human-initiated, then machine-continued texts, class~4 representing deeply-mixed text (where some parts are written by a human and some are generated by a machine), and class~5 representing machine-written, then human-edited texts.

In this notebook, we describe our mdok approach addressing both subtasks (two separate systems). We build on our previously developed robust LLM fine-tuning for sequence classification~\cite{macko2025increasingrobustnessfinetunedmultilingual}, but adjust the training process to the shared task training data and binary/multiclass problem formulation. We further explore the usage of the most recent LLMs, size of which ranging from 1B to 14B parameters. The submitted systems are based on Qwen3-4B and Qwen3-14B models~\cite{qwen3technicalreport}. For the replication purpose, we publish the source code of the mdok approach\footnote{\url{https://github.com/kinit-sk/mdok}}. 

The contributions of the proposed approach are as follows:
\begin{itemize}
    \item We have \textbf{proposed the usage of most modern Qwen3 LLM in robust mdok fine-tuning pipeline and evaluated} its robustness in sensitivity to obfuscation.
    \item We have \textbf{proposed a modification of robust fine-tuning mdok approach for multiclass} detection of AI-human collaboration.
    \item We have \textbf{benchmarked multiple most modern LLMs} on the provided evaluation datasets, making comparison of other models and approaches straightforward and fair.
\end{itemize}

\section{Background}

A binary machine-generated text detection is a well researched task, typically addressed by stylometric methods (e.g., a machine learning classifier trained on TF-IDF features), statistical methods (e.g., utilizing perplexity, entropy, or likelihood)~\cite{hans2024spottingllmsbinocularszeroshot,bao2023fast}, or fine-tuned language models for classification task (e.g., by supervised or contrastive learning)~\cite{spiegel-macko-2024-kinit,dipta2024husemeval2024task8a}. Most of the detection methods can be directly applied by existing frameworks, such as MGTBench~\cite{10.1145/3658644.3670344} or IMGTB~\cite{spiegel-macko-2024-imgtb}.

A multiclass machine-generated text classification is mostly researched in related authorship attribution task, identifying the author (generator) of the text. The problem of different classes of AI-human collaboration can be approached by existing authorship attribution methods~\cite{uchendu-etal-2021-turingbench-benchmark,la2024contrasting}, only slightly different from binary (two-class) methods.

The robustness of machine-generated text detection methods against authorship obfuscation methods has been explored in~\cite{macko-etal-2024-authorship}. Although it has been focused on multilingual settings and the results differ among languages, it has shown that fine-tuned detection methods are more robust against obfuscation than the statistical methods, while offering significantly higher detection performance. Further it has shown that including obfuscated data into fine-tuning process increases the detector's robustness against obfuscation. Such approach has been proposed in~\cite{macko2025increasingrobustnessfinetunedmultilingual} and shown to increase even the generalization to out-of-distribution data. The training data mixture consists of social-media texts from the MultiSocial~\cite{multisocial} dataset, news articles from the MULTITuDE~\cite{macko-etal-2023-multitude} dataset, and obfuscated texts from~\cite{macko-etal-2024-authorship}. For validation (i.e., model-checkpoint selection), the approach uses a unique massively multi-generator (75 generators) multilingual (7 languages) data of MIX2k composition of 18 existing labeled datasets to represent out-of-distribution data.

\section{System Overview}

For this shared task, we aimed to keep the system as simple as possible, ideally resulting in a single-model detection system for each task (avoiding ensembles), just by tweaking the data and the training process.

\begin{figure}
\centering
\includegraphics[width=0.8\textwidth,trim={0 0.5cm 0 0.5cm},clip]{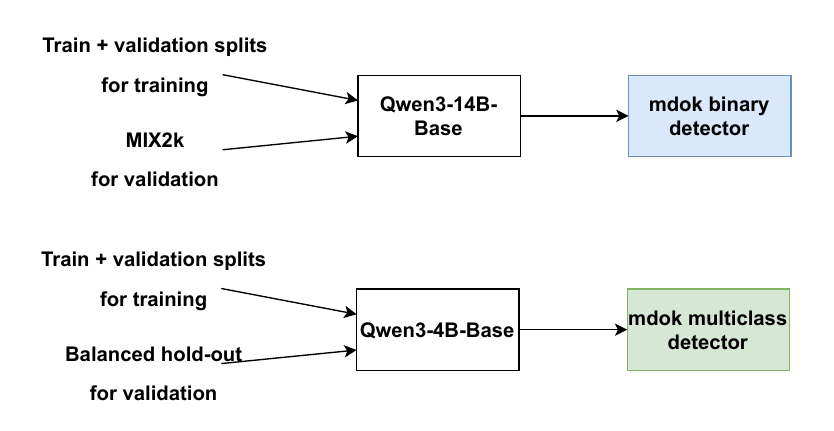}
\caption{An overview of mdok approach for binary (top) and multiclass (bottom) detection of machine-generated text.} \label{overview}
\end{figure}

Although Subtask~1 does not explicitly mention any limitation in usage of additional training data, the Subtask~2 conditions are clear in this regard, not allowing additional training data. Therefore, for adoption of robust fine-tuning approach of~\cite{macko2025increasingrobustnessfinetunedmultilingual}, we do not include additional training data. Instead, for Subtask~1, we combine train and validation sets into training data and modify a small portion of them by using a generic homoglyph attack of~\cite{macko-etal-2024-authorship}. For validation (model selection), we use MIX2k dataset only to ensure the best generalization to out-of-distribution data. For Subtask~2, we also combine train and validation sets for training, but we leave a pseudo-randomly balanced (500 samples per class) holdout portion for validation (in order to minimize bias due to imbalanced evaluation). An overview of this approach is illustrated in Figure~\ref{overview}.

For the fine-tuning process, we have used the adjusted published script for QLoRA based robust fine-tuning of~\cite{macko2025increasingrobustnessfinetunedmultilingual}. Besides of modification of training data mixture, we have used learning rate of $2e-4$, avoided gradient accumulation, and limited the training time to 3 epochs. Due to already incorporated weighted cross entropy for loss calculations, it was straightforward to extend it to multiclass classification by calculating weights based on the training data class distribution.

Since the evaluation on the original validation sets is not usable (due to data leakage) when combined in the training, we have used the original training sets for training to select suitable LLMs to robustly fine-tune for the two subtasks and compared them on validation sets (as well as on MIX2k data). Based on such comparison, we have decided to use the Qwen3-14B model for binary detection in Subtask~1 and the Qwen3-4B for multiclass classification in Subtask~2.

\subsection{Homoglyph Attack}
As mentioned, we have used a generic homoglyph attack of~\cite{macko-etal-2024-authorship} to modify (in the pre-processing step) a small portion of training data (pseudo-randomly selected 10\% of machine-generated texts) for Subtask~1 to increase its robustness to obfuscation. It uses the whole confusables table\footnote{\scriptsize\url{https://www.unicode.org/Public/security/8.0.0/confusables.txt}} to psudorandomly replace letters for their homoglyphs. The probability of a character to be replaced was set to 0.05 (i.e., about 5\% of characters). It further integrates a pseudorandom insertion of visually unseenable zero-width-joiner character in the text (also with a probability of 0.05). We have used the random seed of 42. It has been previously shown that seeing such obfuscated data in the training (fine-tuning) effectively eliminates their highly negative effect on the detection.

\subsection{MIX2k Dataset}
For validation in out-of-distribution settings in Subtask~1, we have used MIX2k dataset, introduced by~\cite{macko2025increasingrobustnessfinetunedmultilingual}. It contains 1,000 samples for each class (human and machine), pseudo-randomly sampled from 18 existing labeled datasets. Such data composition consists of data from 75 generators in 7 languages, thus being mostly out-of-distribution (in comparison to training data). Validation on such a dataset during fine-tuning procedure helps to select the model checkpoint, which has the best generalization to out-of-distribution data, effectively avoiding overfit to training data.

\section{Results}

The results are provided by using standard/official evaluation metrics specified for the shared task. Namely, in Subtask~1, the following metrics are used:
\begin{itemize}
\item \textbf{ROC-AUC}: The area under the ROC (Receiver Operating Characteristic) curve.
\item \textbf{Brier}: The complement of the Brier score (mean squared loss).
\item \textbf{C@1}: A modified accuracy score that assigns non-answers (score = 0.5) the average accuracy of the remaining cases.
\item \textbf{F1}: The harmonic mean of precision and recall.
\item \textbf{F0.5u}: A modified F0.5 measure (precision-weighted F measure) that treats non-answers (score = 0.5) as false negatives.
\item The arithmetic \textbf{mean} of all the metrics above.
\end{itemize}
For Subtask~2, standard metrics utilize \textbf{macro} averages of \textbf{Recall}, \textbf{Precision}, and \textbf{F1} scores, as well as overall \textbf{Accuracy}.

In Table~\ref{subtask1val}, the detection performance for official validation set of Subtask~1 for each tested detector is provided. The results indicate that all the fine-tuned models perform almost ideally (very close to 1 values, outperforming all the baselines), which also indicates a potential overfit to in-distribution data.

\begin{table}
\caption{Comparison of Various Models and Variants on Validation Set of Subtask~1}\label{subtask1val}
\begin{tabular}{l|cccccc}
\toprule
\textbf{Detector} & \textbf{ROC-AUC} & \textbf{Brier} & \textbf{C@1} & \textbf{F1} & \textbf{F0.5u} & \textbf{Mean} \\
\midrule
TF-IDF baseline & 0.996 & 0.951 & 0.984 & 0.980 & 0.981 & 0.978 \\
PPMD baseline & 0.786 & 0.799 & 0.757 & 0.812 & 0.778 & 0.786 \\
Binoculars baseline & 0.918 & 0.867 & 0.844 & 0.872 & 0.882 & 0.877 \\
\midrule
gemma-2-2b & 0.9976 & 0.9980 & 0.9980 & 0.9985 & 0.9981 & 0.9981 \\
gemma-2-9b-it & \bfseries 0.9996 & \bfseries 0.9994 & \bfseries 0.9994 & \bfseries 0.9996 & \bfseries 0.9998 & \bfseries 0.9996 \\
Qwen3-4B-Base & 0.9991 & 0.9989 & 0.9989 & 0.9991 & 0.9997 & 0.9991 \\
Qwen3-8B-Base & \bfseries 0.9996 & \bfseries 0.9994 & \bfseries 0.9994 & \bfseries 0.9996 & \bfseries 0.9998 & \bfseries 0.9996 \\
Qwen3-14B-Base & 0.9994 & 0.9992 & 0.9992 & 0.9994 & 0.9997 & 0.9994 \\
\textbf{mdok (binary)} & 0.9972 & 0.9978 & 0.9978 & 0.9983 & 0.9978 & 0.9978 \\
\bottomrule
\end{tabular}
\end{table}

Therefore, we have also tested the detectors on the MIX2k out-of-distribution dataset (see Table~\ref{subtask1mix}). As the results show, the Qwen3-14B-Base fine-tuned detector generalizes the best to out-of-distribution data. Therefore, it was our natural selection for robust fine-tuning using the mdok approach (described in Section~3), resulting in the mdok (binary) detector, which further boosted the performance, outperforming the baseline detectors.

\begin{table}
\caption{Comparison of Various Models and Variants on MIX2k Out-of-distribution Dataset}\label{subtask1mix}
\begin{tabular}{l|cccccc}
\toprule
\textbf{Detector} & \textbf{ROC-AUC} & \textbf{Brier} & \textbf{C@1} & \textbf{F1} & \textbf{F0.5u} & \textbf{Mean} \\
\midrule
TF-IDF baseline & 0.5342 & 0.6817 & 0.5500 & 0.5837 & 0.5586 & 0.5817 \\
PPMD baseline & 0.4838 & 0.6970 & 0.5395 & 0.2614 & 0.4100 & 0.4783 \\
Binoculars baseline & 0.6368 & \bfseries 0.7443 & 0.6440 & 0.5346 & 0.6554 & 0.6430 \\
\midrule
gemma-2-2b & 0.6230 & 0.6230 & 0.6230 & 0.4512 & 0.6210 & 0.5882 \\
gemma-2-9b-it & 0.5920 & 0.5920 & 0.5920 & 0.3655 & 0.5480 & 0.5379 \\
Qwen3-4B-Base & 0.5710 & 0.5710 & 0.5710 & 0.3500 & 0.5066 & 0.5139 \\
Qwen3-8B-Base & 0.5720 & 0.5720 & 0.5720 & 0.2902 & 0.4797 & 0.4972 \\
Qwen3-14B-Base & 0.6480 & 0.6480 & 0.6480 & 0.5539 & 0.6597 & 0.6315 \\
\textbf{mdok (binary)} & \bfseries 0.6995 & 0.6995 & \bfseries 0.6995 & \bfseries 0.6696 & \bfseries 0.7121 & \bfseries 0.6960 \\
\bottomrule
\end{tabular}
\end{table}

For Subtask~2, the results for the official validation set are provided in Table~\ref{subtask2val}. Although the results indicate that the Qwen2.5-1.5B models performs the best in official metric of Macro Recall, we selected newer and bigger Qwen3-4B-Base (with similar performance) for robust fine-tuning. Using the mdok approach for the fine-tuning resulted in mdok (multiclass) detector, hopefully outperforming the other variants. However, since it uses combined training and validation sets for training, the results indicated in Table~\ref{subtask2val} for this detector are not representative.

\begin{table}
\caption{Comparison of Various Models and Variants on Validation Set of Subtask~2}\label{subtask2val}
\begin{tabular}{l|cccc}
\toprule
\textbf{Detector} & \textbf{Macro Recall} & \textbf{Macro Precision} & \textbf{Macro F1} & \textbf{Accuracy} \\
\midrule
roberta-base baseline & 0.7006 & 0.6515 & 0.6039 & 0.5593 \\
\midrule
Llama-3.2-1B & 0.6172 & 0.6204 & 0.5558 & 0.5385 \\
Qwen2.5-1.5B & 0.7676 & 0.6466 & 0.6325 & 0.5848 \\
Qwen3-1.7B-Base & 0.6348 & 0.6763 & 0.5892 & 0.5466 \\
Qwen2.5-3B & 0.6911 & 0.6072 & 0.5768 & 0.6099 \\
Qwen3-4B-Base & 0.7345 & 0.6267 & 0.6311 & 0.6387 \\
gemma-2-9b-it & 0.6812 & 0.5634 & 0.5609 & 0.5826 \\
\textbf{mdok (multiclass)} & \bfseries 0.9727 & \bfseries 0.9523 & \bfseries 0.9617 & \bfseries 0.9934 \\
\bottomrule
\end{tabular}
\end{table}

\subsection{Subtask~1 Official Results}

The official results are provided in Table~\ref{subtask1leaderboard}, where Mean of the official metrics has been used for ranking. The results show that our approach outperforms all the others in all metrics except of ROC-AUC, where the system of one team outperformed our detector. The performances of the systems are close enough to yield a more efficient detection system, since we have relied on a rather large (14B of parameters) model, consuming more resources for inference. However, the mdok single-model detection system can still be more efficient than some more complex ensemble systems, relying on multiple inference rounds of multiple (although smaller) models. If the labeled test set will be released, an ablation study could pinpoint the strong parts of our system that make it the best performing and/or identify a smaller model to be used instead to provide a more efficient solution.

\begin{table}
\caption{Official Leaderboard of Subtask~1 (for Teams Outperforming the Best-performing Baseline)}\label{subtask1leaderboard}
\begin{tabular}{cl|cccccc}
\toprule
\textbf{Rank} & \textbf{Team Name} & \textbf{ROC-AUC} & \textbf{Brier} & \textbf{C@1} & \textbf{F1} & \textbf{F0.5u} & \textbf{Mean} \\
\midrule
1	&\textbf{mdok (binary)}	&0.853	&\textbf{0.896}	&\textbf{0.894}	&\textbf{0.898}	&\textbf{0.903}	&\textbf{0.899}\\
2	&steely	&0.842	&0.879	&0.877	&0.865	&0.881	&0.880\\
3	&nexus-interrogators	&\textbf{0.865}	&0.874	&0.870	&0.860	&0.881	&0.879\\
4	&yangjlg	&0.845	&0.878	&0.871	&0.856	&0.881	&0.877\\
5	&cnlp-nits-pp	&0.825	&0.873	&0.873	&0.854	&0.882	&0.874\\
6	&unibuc-nlp	&0.828	&0.885	&0.864	&0.845	&0.876	&0.872\\
7	&moadmoad	&0.822	&0.866	&0.865	&0.855	&0.882	&0.871\\
8	&iimasnlp	&0.838	&0.868	&0.856	&0.851	&0.877	&0.869\\
9	&bohan-li	&0.848	&0.858	&0.852	&0.847	&0.870	&0.866\\
10	&advacheck	&0.802	&0.855	&0.855	&0.854	&0.879	&0.863\\
11	&hello-world	&0.838	&0.871	&0.836	&0.827	&0.862	&0.856\\
\bottomrule
\end{tabular}
\end{table}

\subsection{Subtask~2 Official Results}

The official results are provided in Table~\ref{subtask2leaderboard}, where Macro Recall is the official ranking metric. The results show that our approach outperforms all the others by a margin of 3\%. Our decision to keep the system simple and focus on the training data and fine-tuning process paid off. If the labeled test set will be released, the further analysis and ablation study reveals whether we have made the right selection of the system to submit. However, even the top performance of our system (Macro Recall of 64.46\%) is far from perfect (Macro Recall of 100\%). This provides further space for improvements.

\begin{table}
\caption{Official Leaderboard of Subtask~2 (for Teams Outperforming the Baseline)}\label{subtask2leaderboard}
\begin{tabular}{cl|ccc}
\toprule
\textbf{Rank} & \textbf{Team Name} & \textbf{Macro Recall} & \textbf{Macro F1} & \textbf{Accuracy} \\
\midrule
1&	\textbf{mdok (multiclass)}&	\textbf{64.46\%}&	\textbf{65.06\%}&	\textbf{74.09\%} \\
2&	lbh-1130&	61.72\%&	61.73\%&	69.28\% \\
3&	anastasiya.vozniuk&	60.16\%&	60.85\%&	69.04\% \\
4&	Gangandandan&	57.46\%&	56.31\%&	66.81\% \\
5&	Atu&	56.87\%&	56.45\%&	66.30\% \\
6&	TaoLi&	56.74\%&	55.39\%&	66.27\% \\
7&	adugeen&	56.11\%&	55.25\%&	64.79\% \\
8&	Real\_Yuan&	54.49\%&	54.40\%&	62.89\% \\
9&	WeiDongWu&	54.09\%&	53.57\%&	63.01\% \\
10&	zhangzhiliang&	54.06\%&	52.81\%&	61.65\% \\
11&	annepaka22&	54.05\%&	53.49\%&	62.23\% \\
12&	a.dusuki&	52.83\%&	51.44\%&	60.45\% \\
13&	padraig7&	52.14\%&	51.81\%&	59.88\% \\
14&	a.elnenaey&	49.56\%&	50.10\%&	58.96\% \\
15&	Maria Paz&	49.19\%&	48.50\%&	56.94\% \\
\midrule
&   roberta-base baseline&	48.32\%&	47.82\%&	57.09\% \\
\bottomrule
\end{tabular}
\end{table}

\section{Conclusion}

The robustness of machine-generated text detection methods to out-of-distribution data is still challenging. However, we can cope with this problem by better mixture of training data, as we have show in the proposed mdok approach for both binary and multiclass detection. The resulting detectors are very competitive (\textbf{ranking 1st} in both subtasks), outperforming all the baselines, and promising better generalization to out-of-distribution data. Due to gold labels of the test data are not available yet, any ablation study pinpointing crucial parts of the systems and their specific effects was not possible. Further work is still needed to refine the process to achieve ideal performance.

\begin{acknowledgments}
This work was partially supported by the European Union NextGenerationEU through the Recovery and Resilience Plan for Slovakia under the project No. 09I01-03-V04-00059 and partially by \textit{LorAI -- Low Resource Artificial Intelligence}, a project funded by Horizon Europe under \href{https://doi.org/10.3030/101136646}{GA No.101136646}.

\textbf{Computational resources}. We acknowledge EuroHPC Joint Undertaking for awarding us access to Leonardo at CINECA, Italy.
\end{acknowledgments}

\section*{Declaration on Generative AI}
  The author(s) have not employed any Generative AI tools.

\bibliography{pan25-lit,participant-lit}

\end{document}